\documentclass[sigconf,nonacm]{acmart}
\acmDOI{}
\acmISBN{}
\acmPrice{}
\setcopyright{cc}
\renewcommand\footnotetextcopyrightpermission[1]{}
\usepackage{algorithm}
\usepackage{algpseudocode}

\usepackage{censor}
\usepackage{longtable}

\AtBeginDocument{%
  }

\begin{document}

\title[Symbolic Regression with the Beagle System]{GPU-Accelerated  Genetic Programming for Symbolic Regression with Beagle Framework}

\author{Nathan Haut}
\email{hautnath@msu.edu}
\orcid{}
\affiliation{%
  \institution{Michigan State University}
  \streetaddress{}
  \city{East Lansing}
  \state{Michigan}
  \country{USA}
  \postcode{}
}
\author{Ilya Basin}
\email{ilya.basin@noblis.org}
\orcid{}
\affiliation{%
  \institution{Noblis, Inc.}
  \streetaddress{}
  \city{}
  \state{Virginia}
  \country{USA}
  \postcode{}
}
\author{Marzieh Kianinejad}
\email{kianinej@msu.edu}
\orcid{0000-0001-6318-479X}
\affiliation{%
  \institution{Michigan State University}
  \streetaddress{}
  \city{East Lansing}
  \state{Michigan}
  \country{USA}
  \postcode{}
}
\author{Ruchika Gupta}
\email{guptaru1@msu.edu}
\orcid{0009-0004-0696-2902}
\affiliation{%
  \institution{Michigan State University}
  \streetaddress{}
  \city{East Lansing}
  \state{Michigan}
  \country{USA}
  \postcode{}
}
\author{Elijah Smith}
\email{smit3805@msu.edu}
\orcid{}
\affiliation{%
  \institution{Michigan State University}
  \streetaddress{}
  \city{East Lansing}
  \state{Michigan}
  \country{USA}
  \postcode{}
}
\author{Zachary Perrico}
\email{perricoz@msu.edu}
\orcid{0009-0007-8583-3077}
\affiliation{%
  \institution{Michigan State University}
  \streetaddress{}
  \city{East Lansing}
  \state{Michigan}
  \country{USA}
  \postcode{}
}

\author{Wolfgang Banzhaf}
\email{banzhafw@msu.edu}
\orcid{0000-0002-6382-3245}
\affiliation{%
  \institution{Michigan State University}
  \streetaddress{}
  \city{East Lansing}
  \state{Michigan}
  \country{USA}
  \postcode{}
}
\renewcommand{\shortauthors}{Haut et al.}

\begin{abstract}
Beagle is a new software framework that enables execution of Genetic Programming tasks on the GPU. Currently available for symbolic regression, it processes individuals of the population and fitness cases for training in a way that maximizes throughput on extant GPU platforms. In this contribution, we report on the benchmarking of Beagle on the Feynman Symbolic Regression dataset and compare its performance with a fast CPU system called StackGP and the widely available PySR system under the same wall clock budget. We also report on the use of two different fitness functions, one a point-to-point error function, the other a correlation fitness function. The results demonstrate that the Beagle's GPU-aided Symbolic Regression significantly outperforms leading CPU-based frameworks. 
\end{abstract}





\maketitle

\section{Introduction}
Genetic Programming (GP) is a type of evolutionary algorithm that mimics natural evolution by operating on a population of programs - in each generation, GP selects the most fit individuals according to a predefined fitness function using the crossover and mutation genetic operators \cite{koza1992genetic}. Naturally, GP requires a large amount of computational effort as the repeated cycles of evaluation, variation and selection run their course \cite{christensen2002analysis,niehaus2003more}.

Symbolic Regression (SR), a typical application of GP with the goal of finding a mathematical expression to describe data \cite{bnkf}, is a clean laboratory to examine the effects of improvements on efficiency in GP \cite{gustafson2005improving,virgolin2021improving}. SR has broad applications, from nuclear physics to materials science, and is often favored over other ML techniques where interpretability is desired \cite{makke2024interpretable}. 

The evolution of machine learning is inherently linked to the architectural advancements of Graphics Processing Units (GPUs) \cite{oh2004gpu, steinkrau2005using, pandey2022transformational} . GPUs are devices with multicore architectures and parallel processor units. While traditionally classified under the Single Instruction, Multiple Threads (SIMT) model, modern GPU architectures have moved towards Multiple Instruction, Multiple Data (MIMD) functionality, allowing for independent thread scheduling \cite{nvidia2017volta}.

This lineage of GPUs began with the release of NVIDIA’s GeForce 256 (1999) —the industry’s first dedicated GPU \cite{nvidia1999timeline} —which was designed specifically for real-time graphics.  Early researchers leveraged these capabilities by "casting" general mathematical calculations as graphics-specific shaders \cite{nickolls2010gpu}. 

However, the real driving force behind the General-Purpose computing on GPUs (GPGPU) movement was the introduction of Compute Unified Device Architecture (CUDA) in 2007. By offering a C-like programming model, CUDA eliminated the need to map data onto graphics primitives, granting direct access to the GPU instruction set and parallel units. This significantly reduced the barrier to scientific computing and accelerated the evolution of GPUs toward double-precision floating-point arithmetic support \cite{nickolls2008scalable}. As a result, researchers gained access to massively parallel computing resources at a fraction of the cost of traditional high-performance clusters \cite{owens2007survey}.

Early work by Steinkrau et al. \cite{steinkrau2005using} and Chellapilla et al. \cite{chellapilla2006high} revealed the feasibility of accelerating neural networks on GPUs, while the seminal study by Raina, Madhavan, and Ng (2009) \cite{raina2009large} showed that GPU-based training could surpass large CPU clusters for deep neural networks. These cumulative advancements ultimately enabled the breakthrough success of AlexNet in 2012 \cite{krizhevsky2012imagenet}, marking the beginning of the modern deep learning era.
 
The impact of these hardware advancements is evident in benchmarks of standard deep learning models. For example, training ResNet-50, a task that required several days on earlier GPU architectures, can now be completed in a matter of hours \cite{atluri2025evolution}.

Beyond deep learning, GPUs have also addressed the long-standing resource-intensive bottleneck in evolutionary computation by enabling large-scale parallelization of fitness evaluations. This capability has been further exploited in complex hybrid heuristics, such as Memetic Algorithms (MAs), where thousands of independent local searches can be executed concurrently on GPU hardware \cite{luong2009parallel, luong2013parallel}.



In one of the first implementations of genetic programming on GPUs, Harding and Banzhaf \cite{harding2007fast} showed that GPUs could perform several hundred times faster than CPUs when evaluating a Cartesian-GP individual if the number of fitness cases was high. They showed further that the evaluation speedup could also apply to GP image processing tasks \cite{harding2008genetic}. Langdon and Banzhaf \cite{langdon2008simd} utilized GPUs to achieve a speedup with Tree-GP, storing full populations of trees directly on the GPU to reduce overhead. Robilliard et al \cite{robilliard2009high} compared two parallelization schemes and showed that the intensity of GPU speedup over a CPU was affected by the presence of certain operators (such as the \textit{if} operator). Chitty \cite{chitty2012fast} further demonstrated that a multi-core CPU can sometimes outperform the speedup effect of a GPU, although it is unclear whether this result could be challenged by modern graphics processing hardware.

Recent work has highlighted the capabilities of the CUDA library. Sathia et al \cite{sathia2021accelerating} showed an average speedup of 40X when comparing their CUDA-accelerated GP variant against several other standard symbolic regression libraries. Truijillo et al \cite{trujillo2022gsgp} introduced the first CUDA implementation of Geometric Semantic Genetic Programming (GSGP-CUDA) showing speedups greater than 1000X. Zhang et al \cite{zhang2023gpu} successfully applied NVIDIA GPUs to accelerate GP-powered feature extraction in binary image classifiers, while Wang et al. \cite{wang2025evogp} demonstrated how they could also be applied to allow population-level parallelization options in Tree-GP. 


In this work, we aim to explore and benchmark a new GP framework, Beagle \cite{beagle}, that is specifically designed to take advantage of CPU and GPU hardware by strategically distributing computing efforts across all available CPU cores and GPUs to accelerate evolutionary search. Our contributions in this work can be summarized as follows: 

\begin{itemize}
    \item Introducing and describing the Beagle framework,
    \item Benchmarking the performance of the Beagle framework using the Feynman Symbolic Regression Benchmark,
    \item Implementing and assessing performance gains when using a correlation-based fitness function evaluated on the GPU,
    \item and comparing the performance of Beagle against PySR \cite{cranmer2023interpretablemachinelearningscience} and StackGP \cite{haut2024active} to see how it stacks against the state-of-the-art in time-constrained settings.
\end{itemize}



\section{Methods} 

In this contribution, we benchmark the performance of the new GPU-based GP framework, Beagle \cite{beagle}. For benchmarking, we utilize the commonly-used Feynman Symbolic Regression Data \cite{data}. This ensures that results can be easily understood and compared to other systems that have used the same benchmark. We also test the performance of two other GP systems, StackGP \cite{haut2024active} and PySR \cite{cranmer2023interpretablemachinelearningscience} to have fair and comparable performance results using the same computing resources and time constraints. 
Here we provide background and implementation details on the different systems used. In 
Section \ref{sec:beagleOverview} we discuss the new Beagle framework in detail. In Sections \ref{sec:stackgp} and \ref{sec:pysr}, we briefly discuss StackGP and PySR and describe how they were used in this work. In Section \ref{sec:benchmark}, we discuss the benchmarking setup used to compare the GP systems.

\subsection{Beagle Framework}\label{sec:beagleOverview}

Beagle is an open-source symbolic regression framework developed by Ilya Basin at Noblis, Inc that is designed to take advantage of the massive computing power of GPUs and is specifically designed to be "friendly" to heterogenous computing environments \cite{beagle}. One of the key advantages of this design is that it makes it possible to utilize population sizes in the millions to effectively explore large search spaces.   

The Beagle source code can be accessed here: 
https://github.com

/Noblis/beagle-v1.x

\subsubsection{C\# and ILGPU}

Beagle uses the ILGPU open-source C\# library. C\# affords faster execution compared to Python, and ILGPU enables programming at the lowest CUDA level (rather than through the use of higher-level GPU libraries). This, in turn, enables the implementation of arbitrary optimization techniques. Beagle’s development stack is cross-platform and can run well on both Linux and Windows operating systems.

\subsubsection{Approaches for GPU Code}

To maximize performance, Beagle detects multiple installed GPUs and distributes the work between them. It can also handle situations where the entire population cannot fit into a GPU memory at once, in which case the GPU portion is executed in multiple smaller batches that can fit. 

In Beagle’s implementation, the GPU handles evolutionary runs and fitness function evaluations while the CPU handles selection, the birth/death loop, and mutations. In its current implementation Beagle does not use crossover. To minimize the overhead of the resulting data exchange between GPU and CPU, which can be significant, Beagle performs multiple fitness case evaluations per individual in a single generation (with a typical batch size of 512 or 1024). For example, a single generation for a population of 1,000,000 individuals at a batch size of 512 would perform 512,000,000 total fitness case evaluations. Following this approach, Beagle never has to send the same individual to a GPU twice. In a single Beagle generation, all genomes in a population are sent to the GPU together, and each is “asked” to execute a batch of fitness cases. When done, the aggregate results of the fitness function for every individual are returned to the CPU for the selection/births/deaths/mutations processes.

When the Beagle GPU kernel is run, each CUDA block represents a single organism (i.e., the number of blocks equals the population size) and each CUDA thread in the block represents a single fitness case for that organism. This reduces, and often completely eliminates, thread diversion on the GPU, thus avoiding performance degradation related to GPU warp thread diversion. Due to NVIDIA GPU limitations, 1024 is the maximum number of threads per block, meaning 1024 is the maximum number of fitness cases per individual per generation in Beagle.

\subsubsection{Approaches for CPU Code}

On the CPU side, Beagle’s implementation almost completely eliminates Garbage Collection (GC), memory fragmentation, and memory allocation/deallocation overheads, which otherwise may consume more than half of the compute power for evolutionary algorithms. Usually, evolutionary algorithms need to constantly allocate and deallocate memory when individuals are born or die. This causes GC to work “overtime” and creates memory fragmentation issues, thus slowing down performance. To combat this effect, Beagle never deallocates any memory for individuals. Instead memory from “dead” individuals is placed into a “dead pool” and recycled whenever memory is needed for a new individual. In general, Beagle never deallocates memory other than for C\# strings, thus minimizing the overhead of GC and memory defragmentation. 

Where possible, all Beagle CPU code is written to take full advantage of concurrent execution using Microsoft’s Parallels library. This concurrent execution is optimized to eliminate thread contention (i.e., no locks), leading to efficient concurrent execution across all available CPU cores. 

\subsubsection{GPU-Friendly Bespoke Language}

Beagle uses a custom-created Genome Computer Language (GCL), which is a Linear Genetic Programming (LGP) language based on Reverse Polish Notation (RPN). Using LGP as opposed to the typical Tree-Based approach makes the GCL more GC- and GPU-friendly because the GCL is not heap-based, resulting in better performance and scalability.

GCL is structured in a way that produces valid genomes for the majority of mutations. In cases where a valid genome is not produced directly, the mutation mechanism detects the issue and performs a random correction operation. Thus, every mutation in Beagle is guaranteed to produce a viable genome.

\subsubsection{Fitness Functions}

Beagle supports any custom user-defined fitness function that compares fitness cases point-by-point with the target result of the training data. In addition, Beagle supports a correlation-based fitness function, which has previously been shown to improve symbolic regression performance \cite{Haut2023,chen2023relieving} based on earlier work by \cite{keijzer2004scaled}. The foundation of the correlation fitness function is shown in Equation \ref{eq:corr},

\begin{equation}
\label{eq:corr}
    r = \frac{\sum_{i=1}^N (y_i - \bar{y}) (\hat{y}_i - \bar{\hat{y}})}{\sqrt{\sum_{i=1}^N (y_i - \bar{y})^2 \times \sum_{i=1}^N (\hat{y}_i - \bar{\hat{y}})^2}}
\end{equation}
where $N$ is the number of data points $i$, $y_i$ is the target output, and $\hat{y}_i$ is the output from the model. 

The $r$ value is then squared. If a model produces or a dataset contains invalid numbers, rather than removing them, Beagle handles those point pairs separately and rewards models that have invalids in the same locations as the target dataset. Thus, the $r$ value is only representative of all point pairs for a model where both the model and the target output are valid numbers. Once we have $r$, we then compute the total score using Equation \ref{eq:score},

\begin{equation}\label{eq:score}
    score=M * r^4 (N-(c_1+c_2))-M (c_1-c_2)
\end{equation}
where $N$ is the total number of fitness cases, $M$ is a max score parameter set within Beagle, $c_1$ is the number of valid/invalid pairs and $c_2$ is the number of invalid/invalid pairs.

In \cite{Haut2023}, evolution was performed by minimizing $1-r^2$; in this work, we maximize $r^4$. We choose to square  $r^2$  so that we get a wider variance across the very good solutions. Given the specific ranking and selection algorithm in Beagle, this helps prevent evolution from getting stuck at very good but not perfect models. 

The default fitness function in Beagle performs a case-by-case comparison, similar to RMSE, and is defined in Equation \ref{eq:BeagleFit} where both the model output and target output are valid numbers,


\begin{equation}\label{eq:BeagleFit}
\text{Fitness}(\hat{y_i},y_i)
=
{\small
\begin{cases}
M,
& \max(|\hat{y_i}|,|y_i|) = 0, \\[0.8em]

\operatorname{round}\!\left(
M \cdot 
\frac{\min(|\hat{y_i}|,|y_i|)}{\max(|\hat{y_i}|,|y_i|)}
- \frac{M}{11}
\right),
& \hat{y_i}y_i < 0, \\[1.2em]

\operatorname{round}\!\left(
M \cdot 
\frac{\min(|\hat{y_i}|,|y_i|)}{\max(|\hat{y_i}|,|y_i|)}
\right),
& \hat{y_i}y_i \ge 0 .
\end{cases}
}
\end{equation}

where $\hat{y_i}$ is the model output, $y$ is the target output, $M$ is a max fitness value allowed per case, and $\frac{M}{11}$ is a penalty applied if the signs of $\hat{y}$ and $y$ are different, otherwise that term is 0. For the point-to-point comparisons where at least one of the values is invalid (NaNs), $M$ is returned when they are both invalid (to promote agreement), and $-M$ is returned when they disagree (to penalize disagreement). Each of these fitness evaluations is performed in parallel and the results are aggregated to get an overall fitness score. 

For some functions $y = f(x)$, $y$ does not exist on the entire domain from minus infinity to plus infinity. For such functions, Beagle supports symbolic regression through fitness functions that are able to distinguish between a real number result and a non-real number result and react appropriately. Rather than throwing out the cases where the training data contains invalid values (i.e., NaN/PositiveInfinity/ NegativeInfinity) or killing off models that produce such values, Beagle rewards individuals if they produce invalid values in fitness cases where such invalid values exist in the training data. This allows Beagle to leverage potentially useful information that is often discarded in other symbolic regression systems. 

\subsubsection{Population Size}

The combination of the above-described features  enable Beagle to scale to population sizes in the millions. For example, on a relatively modest GPU server with dual Xeon Gold 6426Y and dual NVIDIA A16 chips, a generation typically takes under 0.5 seconds for a population of one million individuals using 512 fitness cases per generation. In practice, Beagle can run well even on high-end laptops with NVIDIA chips.

Once populations reach into the hundreds of thousands – and Beagle often deals with populations in the tens of millions – the ranking-based selection process starts to represent a significant bottleneck. To combat that, Beagle uses a novel high-performance Monte-Carlo-inspired population control approach that mimics the ranking approach but scales better. At a high level, this means that, rather than sorting the entire population by fitness – an approach that does not scale well for large populations – Beagle sorts a random sample of 100 individuals from the population and uses this “population microcosm” to determine percentiles for the fitness distribution for the entire population. These percentiles are then used in a single-pass selection process. Although this “microcosm” approach may be somewhat imprecise, errors cancel each other out over multiple generations. This enables Beagle to efficiently maintain a target population size for large populations. For example, for a population of 1,000,000, Beagle’s approach is approximately 30,000 times faster than the traditional ranking approach.

In this work, we use a variable population size where in the first several generations we start off with a large population of 5 million and the population size gets smaller after a set number of generations. Two different population strategies are used depending on the fitness function being applied. Smaller population sizes are used with the correlation approach since the correlation approach is more costly to run on the GPU than the point-to-point fitness function. This strategy is discussed in more detail in Section \ref{sec:beagleParams}. 
This helps promote diversity exploration early in evolution and then increases the pressure when the population size is reduced. 

\subsubsection{Parameters}\label{sec:beagleParams}

When using Beagle, the parameters in Table \ref{table_beagle_params} were applied. Two dynamic population size strategies were explored. The first population size approach was used with the point-to-point fitness function and the second one was used with the correlation fitness function. The correlation fitness function progresses toward a much smaller population size to balance the fact that computation of the correlation fitness function is more expensive. 

\begin{table}[H]
\centering
\caption{Listed are all the non-default parameters set for Beagle in this work.}
\label{table_beagle_params}
    \resizebox{\columnwidth}{!}{
    \begin{tabular}{|l|p{4cm}|}
    \hline
    Parameter & Setting \\ \hline
    Pt-Pt Pop Size  & 5 mil. (gen <20) 
    
    1 mil. (gen >= 20)  \\ \hline
    Corr Pop Size & 5 mil. (gen < 5)

    3 mil (gen < 10)
    
    1 mil (gen < 15)

    500k (gen < 20)

    250k (gen >=20) \\ \hline
    Operators &
    $+,-,*,/,x^2,\sqrt{x}, 1/x$,
    $cos,sin,tan, arccos, arcsin$, 
    $arctan$, $tanh,log, exp$ \\ \hline
    Selection Strategy & Monte-Carlo-Inspired Rank Selection \\ \hline
    \end{tabular}}
\end{table}

\subsection{StackGP}\label{sec:stackgp}
StackGP is a stack-based genetic programming system implemented in Python that has primarily been applied to symbolic regression tasks \cite{haut2024active}. It implements several state of the art techniques, such as the multi-objective Pareto tournament selection \cite{pareto}, and a correlation fitness function \cite{Haut2023}. 

In this work, we set the following parameters when running StackGP, as shown in Table \ref{table_stackgp_params}. Any parameters not listed here, use the default StackGP settings, as can be found here \cite{stackGPDocs}. For the operators, StackGP has two pre-defined sets. In this work, we chose the "allOps" set, which contains the more expansive set. For clarity, we list all operators that are included in this set in the parameter table.

\begin{table}[H]
\centering
\caption{Listed are all the non-default parameters set for StackGP in this work.}
\label{table_stackgp_params}
    \resizebox{\columnwidth}{!}{
    \begin{tabular}{|l|p{4cm}|}
    \hline
    Parameter & Setting \\ \hline
    Pop Size   & 300  \\ \hline
    Tournament Size    & 5  \\ \hline
    Elitism Rate & 10 \\ \hline
    Operators &
    $+,-,*,/,x^2,\sqrt{x}, 1/x$,
    $cos,sin,tan, arccos, arcsin$, 
    $arctan$, $tanh,log, exp$ \\ \hline
    Selection Strategy & Pareto Tournament \\ \hline
    \end{tabular}}
\end{table}

\subsection{PySR}\label{sec:pysr}
PySR is a popular and widely cited symbolic regression system accessible in Python via a scikit-learn compatible API, making it easy to use and accessible to a broad audience \cite{cranmer2023interpretablemachinelearningscience}. While the API is accessible in Python, PySR utilizes a Julia-based symbolic regression engine in the backend. Due to its popularity and accessibility, it is a good target for benchmarking. 

Table \ref{table_pysr_params} shows all the parameters specifically set for PySR in this work. Any other parameters were left undefined so that the default settings would be used. The default fitness function in PySR is the mean-squared error. 

\begin{table}[H]
\centering
\caption{Listed are all the non-default parameters set for PySR in this work.}
\label{table_pysr_params}
    \resizebox{\columnwidth}{!}{
    \begin{tabular}{|l|p{4cm}|}
    \hline
    Parameter & Setting \\ \hline
    Model Selection   & "best"  \\ \hline
    Parallelism    & "multithreading"  \\ \hline
    Elitism Rate & 10 \\ \hline
    Operators & $+,-,*,/,x^2,\sqrt{x}, 1/x$,
    $cos,sin,tan, arccos, arcsin$, 
    $arctan$, $tanh,log, exp$ \\ \hline %
    \end{tabular}}
\end{table}

\subsection{Benchmark Setup}\label{sec:benchmark}

We utilized the Feynman Symbolic Regression Benchmark \cite{data} to explore the performance of the Beagle framework and compared it against StackGP and PySR under identical time constraints. Since the Feynman Symbolic Regression Benchmark consists of real world physics equations and the typical goal of applying SR is to find equations for real world phenomena, this is a good benchmark for testing the ability to discover a broad range of models in 10-minute and 30-minute runs. 

Two time constraints were utilized to get a sense of how efficient each system is with respect to runtime and to see how the performances vary at different time points. In addition, we chose these short time constraints since we believe that a broad adoption of GP approaches can be achieved only if systems are efficient and can provide users with quality results in less than an hour. Requiring users to wait many hours or overnight for "good-enough" results makes GP approaches unappealing relative to other common machine learning approaches. Thus, while reporting benchmarking on longer time constraints is interesting from a theoretical perspective on system reachability, from a practitioner's point of view, benchmarks on short time constraints are more useful.  

For each of the 100 problems, we executed 10 independent runs with each system to gather performance statistics for each benchmark problem.

Model were trained using 512 data points for each problem and a test set of 128 data points was used to evaluate the quality of each model evolved. When evaluated using the test set, any model that never exceeded 0.1\% error for any test point was considered validated. If a model crosses the threshold for even one test case, it is not considered validated. Overall success for each benchmark problem was recorded if at least half of the independent runs identified a model that was validated on the test set. 

Beyond testing each system using two different time constraints (10 and 30 minutes), we also tested Beagle with two different fitness functions. The first fitness function explored was the default point-to-point fitness function in Beagle that acts similarly to RMSE and is defined in Equation \ref{eq:BeagleFit}. The second fitness function explored is a correlation fitness function, which was shown to lead to improved performance in \cite{Haut2023} and is defined in Equation \ref{eq:corr}. Despite the evaluation of the correlation function being slower on the GPU than the default Beagle fitness function, we wanted to see if the improved search efficiency of the correlation fitness function outweighed the decreased speed.

CPU calculations were performed using Intel(R) Xeon(R) Platinum 8260 CPUs (2.40 GHz) and GPU calculations were performed using NVIDIA v100 GPUs. Beagle, being the only system explored that utilizes GPUs, was given access to 4 CPUs and 2 GPUs for each search. PySR, since it can leverage parallel CPUs was given 4 CPUs for each search. StackGP, which currently only supports single threaded search, was given access to just 1 CPU. 
Thus, each GP system was given the maximum resources available given our hardware constraints.


\section{Results}

Across the benchmark problems, each GP setup was compared using the total number of problems typically solved (defined as problems where 50\% or more of the independent runs found solutions that were validated) and the total number of problems solved at least once (best performance across 10 runs). While the typical performance is more useful for providing an expectation of average performance, reporting the best performance observed is also helpful for indicating whether a solution is reasonably reachable in each system. 

The benchmarking results in the paper represent nearly 11,500 compute hours of effort across CPUs and GPUs. 

\begin{figure}[h]
\centering
\includegraphics[width=8cm]{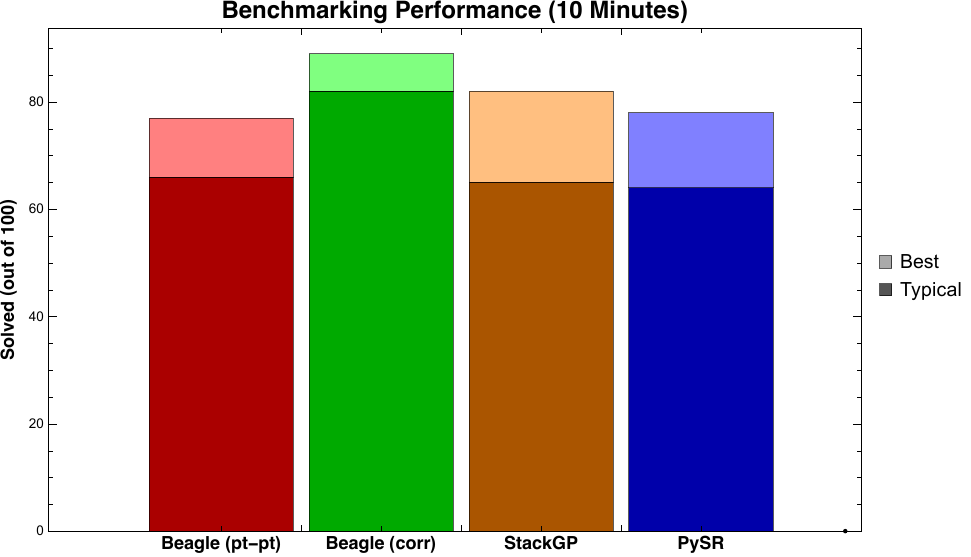}
\caption{The 10 minute benchmarking performance of the different symbolic regression methods. The darker shaded regions indicate the typical performance ($>=50\%$ solved), whereas the lighter region indicates the number of problems solved at least once.}
\label{fig:benchmark10}
\end{figure}

The results of running each system using the 10 minute runtime constraint are shown in Table \ref{table_10min_summary} and Figure \ref{fig:benchmark10}. The column labeled \emph{Beagle (pt-pt)} contains the results of using Beagle with the default point-by-point fitness function. The column labeled \emph{Beagle (corr)} contains the results of using Beagle with the updated correlation fitness function. 
The other two columns contain the results of using StackGP and PySR. The row labeled "typical" indicates the number of problems where the median run found a solution to the problem. The row labeled "best" indicates the number of problems where at least one of the GP runs found a solution to a problem. From the results in the 10-minute runs, we can see that in the typical case, Beagle (corr) achieves the best performance, followed by Beagle(pt-pt), then StackGP, with PySR solving the fewest problems. As well, we confirm that use of the correlation fitness function in Beagle improves the typical performance. Case-by-case results for the 10 minute runs can be found in the Appendix in Table \ref{val10}.

This shows that by effectively utilizing available compute hardware while also using a good fitness function to guide search, we can get significant performance gains in GP. Even with very short time constraints of 10 minutes the system performed well. 

\begin{table}[H]
\centering
\caption{Performance result of each setup in 10 minute runs. Total of typical and best performance reported for each setup across 10 runs. Numbers indicate the number of problems (out of 100) that were typically solved (typical) and solved at least once (best).}
\label{table_10min_summary}
    \resizebox{\columnwidth}{!}{
    \begin{tabular}{|l|c|c|c|c|c|}
    \hline
    Setup & Beagle (pt-pt) & Beagle (corr) & StackGP & PySR \\ \hline
    Typical   & 66 & 82 & 65 & 64 \\ \hline
    Best    & 77 & 89 & 82 & 78 \\ \hline
    \end{tabular}}
\end{table}

The results of running each system using the 30-minute runtime constraint are shown in Table \ref{table_30min_summary} and Figure \ref{fig:benchmark30}. Case-by-case results can be found in the Appendix in Table \ref{val30}. We can see that the extra time allowed each system to solve more problems, indicating that none of the systems has stalled or achieved maximum potential within the 10-minute runs. The results of the typical runs show Beagle (corr) with the best performance, followed by Beagle (pt-pt), then PySR, then StackGP. 

\begin{table}[H]
\centering
\caption{Performance result of each setup in 30 minute runs. Total of typical and best performance reported for each setup across 10 runs. Numbers indicate the number of problems (out of 100) that were typically solved (typical) and solved at least once (best).}
\label{table_30min_summary}
    \resizebox{\columnwidth}{!}{
    \begin{tabular}{|l|c|c|c|c|}
    \hline
    Setup & Beagle (pt-pt) & Beagle (corr) & StackGP & PySR \\ \hline
    Typical   & 74 & 84 & 70 & 72 \\ \hline
    Best    & 80 & 91 & 83 & 86 \\ \hline
    \end{tabular}}
\end{table}

\begin{figure}[h]
\centering
\includegraphics[width=8cm]{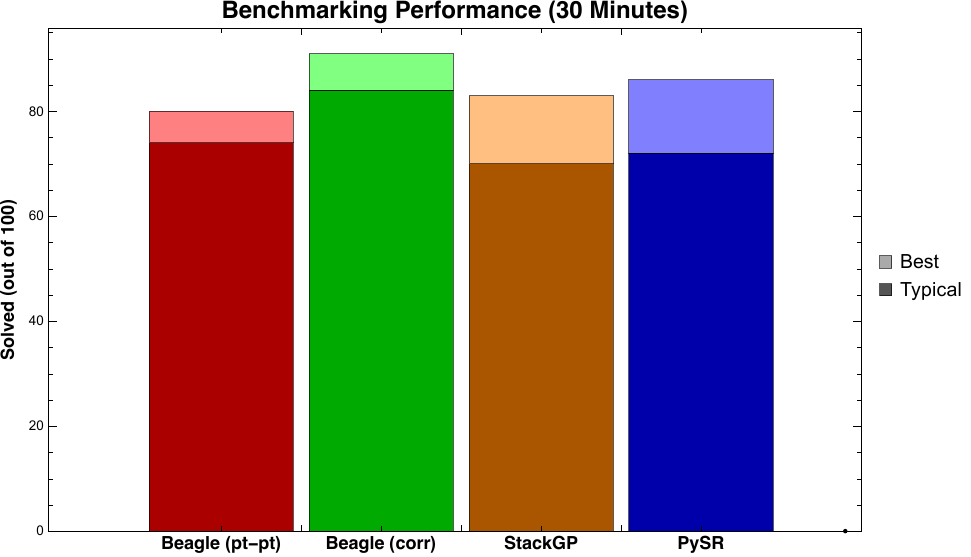}
\caption{The 30 minute benchmarking performance of the different symbolic regression methods. The darker shaded regions indicate the typical performance ($>=50\%$ solved), whereas the lighter region indicates the number of problems solved at least once.}
\label{fig:benchmark30}
\end{figure}

Because Beagle supports fitness functions that handle NaNs, we also explored the impact that incorporating  knowledge about the locations of NaNs into the fitness function has on the success of a search using the Beagle framework. To do this, we investigated Beagle's ability to find the quadratic equation formula in two different settings. The first setting restricted the domain of training data such that only real values would be generated, similar to what was done when generating the Feynman data. The second scenario allowed a wider domain, such that imaginary numbers would be created which are automatically converted to NaNs in Beagle. We ran both cases using a 10-minute runtime constraint and executed 10 independent runs for each. The results show that considering the location of NaNs greatly improved the performance. In the case where we restricted the domain to avoid producing imaginary numbers, Beagle did not find a perfect solution in any of the 10 runs. In the case where we allowed imaginary numbers to be produced, all 10 runs found a perfect solution. Neither StackGP nor PySR support fitness functions that handle NaNs, and neither of these two systems was able to find a perfect solution in any of the 10 runs. These results using information about the location of NaNs in Beagle indicate that if we were to expand the domains used when generating the data for the Feynman problems rather than restrict them to only producing real values, we may be able to solve an even larger set of Feynman problems. Running the full benchmark with wider input domains serves as a great target for future work. 


\section{Conclusion}

In this work, we benchmarked the new open-source Beagle GP framework by utilizing the Feynman Symbolic Regression Benchmark and compared it against two other open-source GP systems, StackGP and PySR. The results show that Beagle, in runtime constrained settings, outperforms both StackGP and PySR, solving a larger set of the benchmark problems in the typical case. These results show that Beagle has successfully mapped GP to GPU hardware to realize significant performance gains. In addition, we determined that we could successfully implement a correlation-based fitness function in Beagle that runs efficiently using GPUs. Although each individual fitness evaluation is a bit slower when using the correlation fitness function compared to the point-by-point fitness function, the improvements in search efficiency when using the correlation fitness function lead to overall improved performance. 

While both StackGP and PySR are competitive SR systems, for users and researchers with access to GPUs, SR performance can be improved using Beagle to access the power of those GPUs. This improved runtime performance achieved by using GPUs could have widespread impacts that make genetic programming more appealing and accessible for broader applications where time-constraints are tight and problems are challenging, requiring efficient exploration of huge search spaces. Furthermore, the ability to efficiently manage population sizes on the order of 1 million or more individuals could open doors for GP researchers, since much greater population diversity and other interesting dynamics could be achieved compared to traditional CPU-bound systems.

\begin{acks}
   The team acknowledges Noblis, Inc. for supporting the development of the Beagle system. This work was supported in part by Michigan State University through computational resources provided by the Institute for Cyber-Enabled Research.
\end{acks}







\section{Appendices}
\appendix

\onecolumn

\begin{longtable}{|l|l|l|l|l|}
\caption{The validation results of models developed in the 10 minute runs using each symbolic regression system. The numbers indicate the number of validated runs out of 10, except problem 7 where we increased to 20 runs to determine if typical performance was statistically significant. }\label{val10}
\\ \hline
EQ \# & Beagle (pt-pt) & Beagle (Corr) & StackGP & PySR \\ \hline
1     & 10             & 10            & 9       & 1    \\ \hline
2     & 0              & 10            & 2       & 0    \\ \hline
3     & 0              & 10            & 0       & 0    \\ \hline
4     & 0              & 10            & 4       & 0    \\ \hline
5     & 0              & 0             & 0       & 0    \\ \hline
6     & 10             & 7             & 9       & 7    \\ \hline
7     & 1              & 12 (/20)             & 0       & 10   \\ \hline
8     & 10             & 10            & 10      & 10   \\ \hline
9     & 0              & 10            & 1       & 0    \\ \hline
10    & 10             & 10            & 10      & 9    \\ \hline
11    & 10             & 10            & 10      & 10   \\ \hline
12    & 10             & 10            & 10      & 10   \\ \hline
13    & 10             & 10            & 10      & 10   \\ \hline
14    & 4              & 2             & 5       & 8    \\ \hline
15    & 10             & 10            & 10      & 10   \\ \hline
16    & 10             & 10            & 10      & 10   \\ \hline
17    & 8              & 10            & 2       & 0    \\ \hline
18    & 0              & 0             & 0       & 0    \\ \hline
19    & 10             & 10            & 3       & 4    \\ \hline
20    & 0              & 0             & 0       & 0    \\ \hline
21    & 0              & 0             & 0       & 5    \\ \hline
22    & 10             & 10            & 10      & 10   \\ \hline
23    & 10             & 10            & 10      & 9    \\ \hline
24    & 0              & 10            & 1       & 1    \\ \hline
25    & 10             & 10            & 10      & 10   \\ \hline
26    & 10             & 5             & 10      & 10   \\ \hline
27    & 10             & 10            & 3       & 3    \\ \hline
28    & 10             & 10            & 10      & 10   \\ \hline
29    & 0              & 0             & 0       & 0    \\ \hline
30    & 1              & 6             & 2       & 0    \\ \hline
31    & 10             & 4             & 10      & 10   \\ \hline
32    & 10             & 10            & 10      & 5    \\ \hline
33    & 0              & 10            & 5       & 0    \\ \hline
34    & 10             & 10            & 10      & 10   \\ \hline
35    & 10             & 10            & 3       & 8    \\ \hline
36    & 4              & 9             & 2       & 4    \\ \hline
37    & 10             & 10            & 10      & 10   \\ \hline
38    & 0              & 0             & 0       & 0    \\ \hline
39    & 10             & 10            & 10      & 9    \\ \hline
40    & 10             & 10            & 10      & 10   \\ \hline
41    & 10             & 10            & 10      & 10   \\ \hline
42    & 10             & 10            & 10      & 10   \\ \hline
43    & 1              & 0             & 0       & 0    \\ \hline
44    & 0              & 0             & 0       & 0    \\ \hline
45    & 10             & 10            & 10      & 10   \\ \hline
46    & 10             & 10            & 10      & 10   \\ \hline
47    & 10             & 10            & 6       & 10   \\ \hline
48    & 10             & 10            & 10      & 3    \\ \hline
49    & 10             & 10            & 10      & 10   \\ \hline
50    & 4              & 7             & 0       & 3    \\ \hline
51    & 1              & 10            & 1       & 3    \\ \hline
52    & 10             & 10            & 10      & 10   \\ \hline
53    & 10             & 10            & 10      & 10   \\ \hline
54    & 10             & 10            & 10      & 10   \\ \hline
55    & 10             & 10            & 10      & 7    \\ \hline
56    & 0              & 2             & 0       & 0    \\ \hline
57    & 5              & 1             & 9       & 1    \\ \hline
58    & 10             & 10            & 10      & 9    \\ \hline
59    & 10             & 10            & 10      & 10   \\ \hline
60    & 10             & 10            & 10      & 10   \\ \hline
61    & 10             & 10            & 5       & 3    \\ \hline
62    & 0              & 10            & 9       & 6    \\ \hline
63    & 10             & 10            & 10      & 9    \\ \hline
64    & 0              & 10            & 0       & 4    \\ \hline
65    & 1              & 10            & 10      & 9    \\ \hline
66    & 10             & 10            & 10      & 9    \\ \hline
67    & 10             & 10            & 5       & 6    \\ \hline
68    & 10             & 10            & 6       & 6    \\ \hline
69    & 10             & 10            & 10      & 10   \\ \hline
70    & 10             & 10            & 10      & 10   \\ \hline
71    & 0              & 10            & 1       & 0    \\ \hline
72    & 0              & 0             & 0       & 2    \\ \hline
73    & 10             & 10            & 10      & 10   \\ \hline
74    & 10             & 10            & 10      & 10   \\ \hline
75    & 10             & 10            & 10      & 10   \\ \hline
76    & 10             & 10            & 10      & 10   \\ \hline
77    & 10             & 10            & 10      & 10   \\ \hline
78    & 10             & 10            & 10      & 10   \\ \hline
79    & 10             & 10            & 10      & 10   \\ \hline
80    & 0              & 10            & 1       & 0    \\ \hline
81    & 4              & 10            & 0       & 0    \\ \hline
82    & 2              & 10            & 2       & 4    \\ \hline
83    & 10             & 10            & 10      & 10   \\ \hline
84    & 10             & 10            & 10      & 10   \\ \hline
85    & 10             & 10            & 10      & 10   \\ \hline
86    & 0              & 3             & 3       & 0    \\ \hline
87    & 0              & 0             & 0       & 0    \\ \hline
88    & 10             & 10            & 10      & 10   \\ \hline
89    & 10             & 10            & 10      & 7    \\ \hline
90    & 0              & 0             & 0       & 0    \\ \hline
91    & 0              & 3             & 0       & 1    \\ \hline
92    & 10             & 10            & 10      & 10   \\ \hline
93    & 10             & 10            & 10      & 10   \\ \hline
94    & 4              & 10            & 1       & 8    \\ \hline
95    & 7              & 8             & 10      & 9    \\ \hline
96    & 10             & 10            & 10      & 10   \\ \hline
97    & 10             & 10            & 10      & 9    \\ \hline
98    & 10             & 10            & 10      & 10   \\ \hline
99    & 9              & 10            & 4       & 0    \\ \hline
100   & 10             & 10            & 10      & 9    \\ \hline
\end{longtable}

\begin{longtable}{|l | l | l | l | l |}
\caption{The validation results of models developed in the 30 minute runs using each symbolic regression system. The numbers indicate the number of validated runs out of 10.}\label{val30}
\tiny
\\ \hline
EQ \# & Beagle (pt-pt) & Beagle (Corr) & StackGP & PySR \\ \hline
1     & 10             & 10            & 10      & 3    \\ \hline
2     & 0              & 10            & 2       & 1    \\ \hline
3     & 0              & 10            & 0       & 1    \\ \hline
4     & 1              & 10            & 10      & 1    \\ \hline
5     & 0              & 1             & 0       & 0    \\ \hline
6     & 10             & 7             & 10      & 9    \\ \hline
7     & 3              & 9             & 1       & 10   \\ \hline
8     & 10             & 10            & 10      & 10   \\ \hline
9     & 0              & 10            & 1       & 0    \\ \hline
10    & 10             & 10            & 10      & 10   \\ \hline
11    & 10             & 10            & 10      & 10   \\ \hline
12    & 10             & 10            & 10      & 10   \\ \hline
13    & 10             & 10            & 10      & 10   \\ \hline
14    & 10             & 3             & 1       & 6    \\ \hline
15    & 10             & 10            & 10      & 10   \\ \hline
16    & 10             & 10            & 10      & 10   \\ \hline
17    & 10             & 10            & 8       & 0    \\ \hline
18    & 0              & 0             & 0       & 1    \\ \hline
19    & 10             & 10            & 6       & 10   \\ \hline
20    & 0              & 0             & 0       & 2    \\ \hline
21    & 0              & 0             & 0       & 10   \\ \hline
22    & 10             & 10            & 10      & 10   \\ \hline
23    & 10             & 10            & 10      & 10   \\ \hline
24    & 3              & 10            & 2       & 0    \\ \hline
25    & 10             & 10            & 10      & 10   \\ \hline
26    & 10             & 5             & 10      & 10   \\ \hline
27    & 10             & 10            & 6       & 7    \\ \hline
28    & 10             & 10            & 10      & 10   \\ \hline
29    & 0              & 0             & 0       & 0    \\ \hline
30    & 5              & 6             & 4       & 3    \\ \hline
31    & 10             & 4             & 10      & 10   \\ \hline
32    & 10             & 10            & 10      & 8    \\ \hline
33    & 0              & 10            & 4       & 0    \\ \hline
34    & 10             & 10            & 10      & 10   \\ \hline
35    & 10             & 10            & 5       & 10   \\ \hline
36    & 8              & 9             & 1       & 9    \\ \hline
37    & 10             & 10            & 10      & 10   \\ \hline
38    & 4              & 4             & 0       & 0    \\ \hline
39    & 10             & 10            & 10      & 9    \\ \hline
40    & 10             & 10            & 10      & 10   \\ \hline
41    & 10             & 10            & 10      & 10   \\ \hline
42    & 10             & 10            & 10      & 10   \\ \hline
43    & 0              & 0             & 0       & 0    \\ \hline
44    & 0              & 0             & 0       & 0    \\ \hline
45    & 10             & 10            & 10      & 10   \\ \hline
46    & 10             & 10            & 10      & 10   \\ \hline
47    & 10             & 10            & 9       & 10   \\ \hline
48    & 10             & 10            & 10      & 4    \\ \hline
49    & 10             & 10            & 10      & 10   \\ \hline
50    & 9              & 7             & 1       & 7    \\ \hline
51    & 8              & 10            & 0       & 6    \\ \hline
52    & 10             & 10            & 10      & 10   \\ \hline
53    & 10             & 10            & 10      & 10   \\ \hline
54    & 10             & 10            & 10      & 10   \\ \hline
55    & 10             & 10            & 10      & 10   \\ \hline
56    & 0              & 2             & 0       & 0    \\ \hline
57    & 10             & 0             & 10      & 5    \\ \hline
58    & 10             & 10            & 10      & 10   \\ \hline
59    & 10             & 10            & 10      & 10   \\ \hline
60    & 10             & 10            & 10      & 10   \\ \hline
61    & 10             & 10            & 8       & 2    \\ \hline
62    & 0              & 10            & 10      & 10   \\ \hline
63    & 10             & 10            & 10      & 10   \\ \hline
64    & 0              & 10            & 0       & 3    \\ \hline
65    & 2              & 10            & 10      & 10   \\ \hline
66    & 10             & 10            & 10      & 10   \\ \hline
67    & 10             & 10            & 10      & 9    \\ \hline
68    & 10             & 10            & 9       & 8    \\ \hline
69    & 10             & 10            & 10      & 10   \\ \hline
70    & 10             & 10            & 10      & 10   \\ \hline
71    & 0              & 10            & 3       & 2    \\ \hline
72    & 0              & 1             & 1       & 4    \\ \hline
73    & 10             & 10            & 10      & 10   \\ \hline
74    & 10             & 10            & 10      & 10   \\ \hline
75    & 10             & 10            & 10      & 10   \\ \hline
76    & 10             & 10            & 10      & 10   \\ \hline
77    & 10             & 10            & 10      & 10   \\ \hline
78    & 10             & 10            & 10      & 10   \\ \hline
79    & 10             & 10            & 10      & 10   \\ \hline
80    & 0              & 10            & 0       & 1    \\ \hline
81    & 9              & 10            & 0       & 5    \\ \hline
82    & 10             & 10            & 1       & 9    \\ \hline
83    & 10             & 10            & 10      & 10   \\ \hline
84    & 10             & 10            & 10      & 10   \\ \hline
85    & 10             & 10            & 10      & 10   \\ \hline
86    & 0              & 1             & 3       & 0    \\ \hline
87    & 0              & 0             & 0       & 0    \\ \hline
88    & 10             & 10            & 10      & 10   \\ \hline
89    & 10             & 10            & 10      & 9    \\ \hline
90    & 0              & 0             & 0       & 0    \\ \hline
91    & 1              & 9             & 0       & 0    \\ \hline
92    & 10             & 10            & 10      & 10   \\ \hline
93    & 10             & 10            & 10      & 10   \\ \hline
94    & 6              & 10            & 5       & 8    \\ \hline
95    & 10             & 8             & 10      & 10   \\ \hline
96    & 10             & 10            & 10      & 10   \\ \hline
97    & 10             & 10            & 10      & 10   \\ \hline
98    & 10             & 10            & 10      & 10   \\ \hline
99    & 10             & 10            & 5       & 3    \\ \hline
100   & 10             & 10            & 10      & 10   \\ \hline
\end{longtable}

\twocolumn
\bibliographystyle{ACM-Reference-Format}
\bibliography{acmart-primary/Article/bib}
\onecolumn

\end{document}